# IMAGE DEBLURRING BASED ON LIGHTWEIGHT MULTI-INFORMATION FUSION NETWORK

*Yanni Zhang*[1,+] *Yiming Liu*[2,+] *Qiang Li*[1] *Miao Qi*[1] *Dahong Xu*[2] *Jun Kong*[1,*] *Jianzhong Wang*[1,*]

[1] College of Information Science and Technology, Northeast Normal University, China
[2] College of Information Science and Engineering, Hunan Normal University, China

## ABSTRACT

Recently, deep learning based image deblurring has been well developed. However, exploiting the detailed image features in a deep learning framework always requires a mass of parameters, which inevitably makes the network suffer from high computational burden. To solve this problem, we propose a lightweight multi-information fusion network (LMFN) for image deblurring. The proposed LMFN is designed as an encoder-decoder architecture. In the encoding stage, the image feature is reduced to various small-scale spaces for multi-scale information extraction and fusion without a large amount of information loss. Then, a distillation network is used in the decoding stage, which allows the network benefit the most from residual learning while remaining sufficiently lightweight. Meanwhile, an information fusion strategy between distillation modules and feature channels is also carried out by attention mechanism. Through fusing different information in the proposed approach, our network can achieve state-of-the-art image deblurring result with smaller number of parameters and outperforms existing methods in model complexity.

## 1. INTRODUCTION

The movement of object, shake of camera and out-of-focus will cause the blur artifacts in image, which affects people's visual perception on the detailed image information. Therefore, image deblurring has become an indispensable step in image enhancement. However, estimating the blur kernel to restore sharp images from the blurred ones is difficult since it is a highly ill-posed problem. Most traditional methods [1-3] regularized the solution space by introducing some prior information to model the blur kernel. However, because the prior information may not conform to the real scene in which the blur kernel is more complex than modeled, the quality of restored image is not optimal.

With the rapid process of deep learning, many convolutional neural networks (CNNs) based methods [4-12] have been proposed for effective image deblurring. Compared with the traditional shallow models [1-3], these deep CNN methods do not need to estimate blur kernel, but directly predict sharp image from the blurred one. Thus, they have achieved superior image deblurring results. Multi-scale strategy is commonly used in image deblurring task since it could aggregate features in a coarse-to-fine manner [4,6]. In some methods [8-9], image deblurring is treated as a deconvolution problem, and the networks are designed to approximate deconvolution operations for expanding the feature map. Inspired by the generative adversarial network (GAN), many works [10-12] based on conditional GAN were proposed for image deblurring. Moreover, the encoder-decoder based network is also employed for image deblurring [5,7]. However, the number of parameters in the above approaches is large because they need to stack a mass of convolution layers for feature extraction, which inevitably increases the computational cost of the networks.

In order to reduce the computational burden in image deblurring task, a lightweight multi-information fusion network (LMFN) is proposed. Our LMFN is significantly more efficient with less parameters while guaranteeing the deblurring quality.

Our main contributions are three-folds:

- We propose a multi-scale hierarchical information fusion scheme (MSHF) to encode the feature of blurred image. MSHF extracts and fuses the image feature in multiple small-scale spaces, which can better eliminate redundant parameters while maintaining the rich image information.
- We employ a very lightweight distillation network to decode the image feature back to a sharp image. To the best of our knowledge, it is the first time that the distillation network is adopted in image deblurring problem.
- Two attention mechanism based modules are also presented in the decoding process of our approach to exploit the interdependency between the layers and feature channels, so that a better information fusion can be achieved.

## 2. RELATED WORK

**Image Deblurring**. In recent years, CNNs based image deblurring have been developed rapidly. Nah et al. [4] proposed a multi-scale CNN method for image deblurring called DeepDeblur, which is based on a coarse-to-fine strategy to restore the sharp image progressively. Although it achieved satisfied results, DeepDeblur contains 40 convolutional layers in each scale without parameter sharing. Therefore, the large number of parameters makes the network training be time-consuming. Zhang et al. [8] proposed a spatial variant neural network which consists of three CNNs and a recurrent neural network (RNN) for dynamic scene deblurring. Although this algorithm is effective, it uses a large number of convolutional layers for feature extraction and weights estimation for RNNs, which increase the number of parameters. Furthermore, pre-training the RNNs on VGG16 [13] also increases the computational burden of this network. To lighten the computation issue, Tao et al. [5] employed an encoder-decoder structure to propose a scale-recurrent network (SRN). Compared with

+Equal contribution author.
*Corresponding author.

DeepDeblur, SRN applies the long-short term memory (LSTM) to share weights across scales. Thus, it can achieve highly effective deblurring result. Similarly, the parameter sharing scheme is also adopted by Gao et al. [6] to improve the efficiency of their image deblurring network. Besides, an effective multi-patch model is proposed by Zhang et al. [7] for image deblurring via a fine-to-coarse hierarchical representation.

**Distillation Network.** The information distillation network is one of the sate-of-the-art methods to reduce the number of parameters and achieve a lightweight network architecture. Hui et al. [14] proposed an information distillation network (IDN) which reduces the computational complexity and memory consumption by channel splitting strategy to downscale the feature maps. Based on IDN, a fast and lightweight information multi-distillation network (IMDN) [15] was presented. IMDN extracted features at a granular level by applying the channel splitting strategy multiple times and proposed a contrast-aware channel attention (CCA) to connect the extracted features. However, Liu et al. pointed out that IMDB is still inflexible and inefficient [16]. As a result, they introduced the residual feature distillation block (RFDB), which utilizes the feature distillation connection instead of the channel splitting strategy. Thus, RFDB can improve the performance without introducing additional parameters. Although the distillation network has already been employed in some computer vision problems, there is few studies adopt it for image deblurring. Furthermore, the hierarchical information from different distillation layers is also neglected in the existing methods.

**Attention Mechanism.** Inspired by its successful applications in natural language processing, the attention mechanism has been widely used in image processing tasks [17-21]. Zhang et al. [17] leveraged attention mechanism to allow the network to focus on the relationship among spatial image areas. Kuldeep et al. [18] utilized the weighted sum of all location features to selectively aggregate the location feature. Kim et al. [19] learned the correlation between feature channels through residual blocks and spatial channel attention. Dai et al. [20] used second-order feature statistics to adaptively refine features by a second-order channel attention (SOCA) module. Although the above methods take the correlations of spatial and channels into consideration, the interdependency of different network layers is neglected. More recently, Niu et al. [21] proposed a holistic attention network (HAN) to characterize the relationship between network layers. Nevertheless, HAN regards multiple feature channels of each layer as a whole group and only focuses on estimating the correlation between feature channel groups of different layers. Thus, the correlation between the inter-layer feature channels is ignored. Moreover, the 3D convolution in HAN increases number of parameters and computational burden dramatically.

## 3. PROPOSED METHOD

### 3.1. Overview

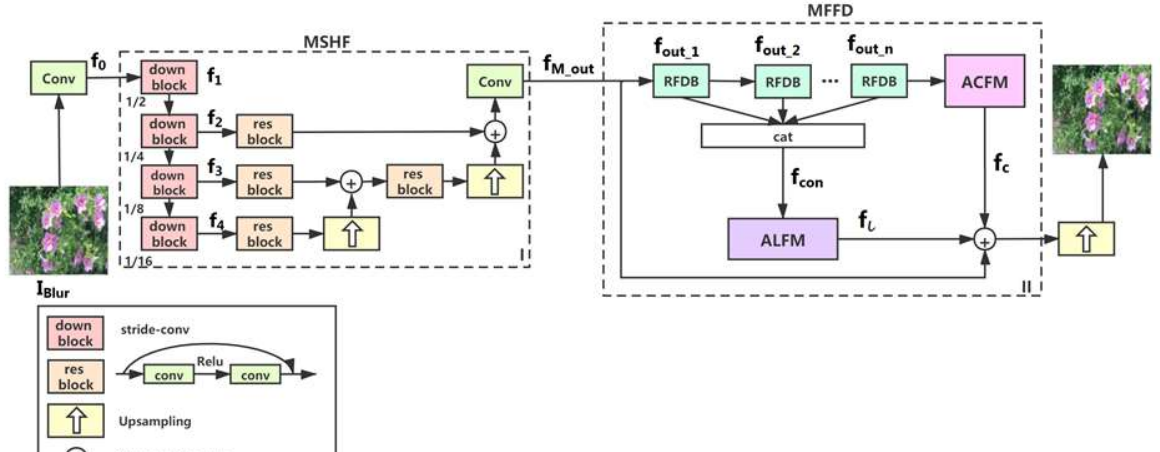

**Fig. 1.** The architecture of lightweight multi-information fusion network (LMFN). Zoom in for better view.

The overall architecture of the proposed LMFN is shown in Fig. 1. As can be seen, it is based on an encoder-decoder structure and consists of two main parts. In Part I, a blurred image after convolution is input into our multi-scale hierarchical fusion module (MSHF) for down-sampling. Through MSHF, the input will be encoded into several small-scale features, and the loss of information will be compensated by the information fusion between different layers. Part II is the multi-feature fusion distillation module based on attention mechanism (MFFD), which will be used for feature extraction and generate a sharp image.

### 3.2. Multi-scale hierarchical fusion module

MSHF is a down-sampling module which serves as the encoder in our approach. Firstly, the shallow feature $f_0$ of an input blurred image $I_{Blur}$ is extracted by a convolutional layer with the kernel size of 3×3 and 64 output channels. Then, $f_0$ is gradually scaled down to four smaller size features $f_i$ (*i*=1,2,3,4) by using downblock modules [22]. After each downblock for down-sampling, we simply use resblock [4] to realize residual learning. In addition, we fuse different scale features in the small-scale space. The small-scale features after residual learning are up-sampled and then element-wise added with the features of adjacent layers. Finally, the more fine-grained feature $f_{M_out}$ in a small-scale can be obtained.

Many existing methods [6,7,18,23,24] mostly carry out multiple complex convolution operations at each layer after down-sampling to prevent the network from losing important detailed information, which will overload the network with large parameters. In our work, we only carry out a small number of feature extraction operations, and then fuse multi-scale hierarchical information in small-scale space. This strategy can effectively reduce the computational cost.

### 3.3. Multi-feature fusion module based on attention mechanism

#### *3.3.1. Distillation block*

After MSHF, our goal is to restore the small-scale features to its original scale and reconstruct a sharp image, which can be done by the decoder process called MFFD. Generally, in order to well reconstruct a sharp image, it is necessary to increase the number of convolutional layers in the network so that the receptive filed can be enlarged to get more information. However, this strategy is not a good choice in practice because it dramatically increases the number of parameters and intermedia feature channels. In the decoding stage, we also pursue a light and fast network. Therefore, the residual feature distillation block (RFDB) [16] is adopted. RFDB makes the network lightweight because it consists of two parts for feature extraction, one is retained and the other is further refined. Furthermore, the shallow residual block in RFDB also makes it benefit from the residual learning.

#### *3.3.2. Fusion mechanism*

Unlike other networks [14-16] that simply stack several distillation modules to hierarchically extract features and only employ final features for the specific task, we rethink the distillation block in our proposed approach. We argue that although the multiple distillation modules (i.e., RFDBs) help the network to obtain richer image information with fewer parameters, the correlation between the intermedia features of each RFDB is ignored. At the same time, we also think about the problem that how to make the information of the last layer fully used. Thus, two different feature fusion mechanisms based on attention are proposed to improve the representation ability of extracted features. One is the attention layer fusion module (ALFM) which is employed to learn the correlation between feature channels obtained by multiple RFDB layers. The other is the attention channel fusion module (ACFM) which is used to describe the dependency between the inter-channel and intra-channel information in adjacent feature channels of the last layer.

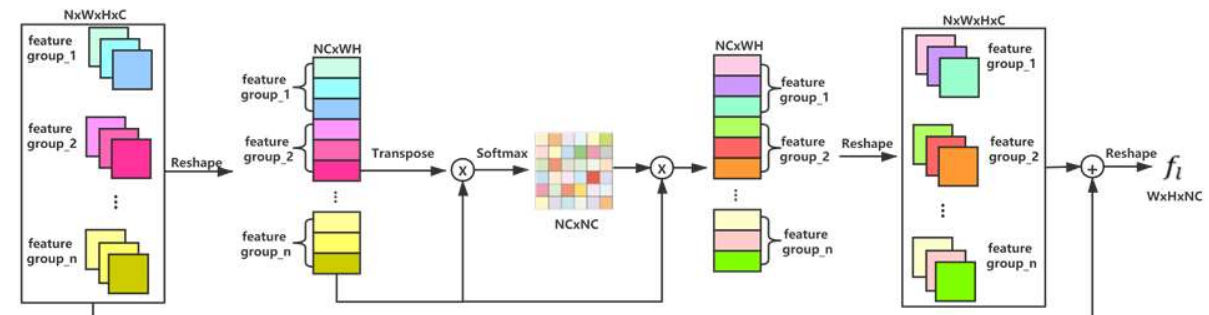

**Fig. 2.** ALFM: the architecture of attention layer fusion module. Zoom in for better view.

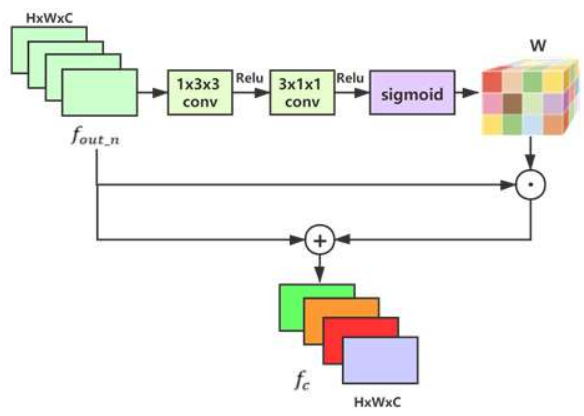

**Fig. 3.** ACFM: the architecture of attention channel fusion module.

The structure of ALFM is shown in Fig. 2. Given the groups of feature channels obtained by *N* RFDB layers with the dimension of *N*×*C*×*W*×*H*, we first reshape them along the channel and get a *NC*×*WH* feature matrix. Then, the feature matrix is transposed and multiplied by itself. After softmax, we can get an attention matrix to reflect the correlation between channels. At last, we multiply the attention matrix by feature matrix with a scale factor $\theta$ and fuse it with the original feature channels by addition operation. The process can be expressed in Equation 1 and 2:

$$m_{ij} = softmax(reshape(f_g) \otimes reshape(f_g)^T) \tag{1}$$

$$f_l = \theta \sum_{i=1}^{N} m_{ij} \otimes f_g + f_g \tag{2}$$

where $f_g$ is the input feature channels, $\otimes$ denotes the matrix multiplication, $m_{ij}$ is the attention matrix, $\theta$ is initialized to 0 and will be automatically optimized by the network. Here, it should be noted that our ALFM treats the feature channels of each RFBD layer separated rather than a whole. Thus, the correlation between feature channels from both the same and different layers can be considered.

The structure of ACFM is shown in Fig. 3. The aim of ACFM is to model the interdependency between feature channels of the last RFBD layer by jointly considering channel and spatial information. Nevertheless, different from HAN [21] which adopted 3D convolution to accomplish this task, our ACFM leverage a Pseudo-3D convolution strategy [25] to reduce the number of parameters. Taking the output of the last RFDB $f_{out_n}$ as input, we first utilize two convolution kernels with the size of 1×3×3 and 3×1×1 to capture the spatial and channel correlations. Then, the attention matrix *W* obtained after sigmoid function is element-wise multiplied by $f_{out_n}$. Finally, the weighted $f_{out_n}$ is scaled by a factor $\alpha$ and fused with the original features.

$$f_c = f_{out_n} + \alpha \cdot \delta(\mathrm{conv2}(\mathrm{conv1}(f_{out_n}))) \odot f_{out_n} \tag{3}$$

where $\delta(\cdot)$ is the sigmoid function, $\odot$ represents element-wise product, conv1 is 1×3×3 convolution, conv2 is 3×1×1 convolution, and α is an optimizable parameter.

Through the attention mechanism in ALFM and ACFM, more powerful feature representation can be achieved, which will compensate the potential information loss in lightweight RFDB.

Overall, the loss function of our network can be expressed by Equation 4:

$$L_{MSE} = \frac{1}{wh} \sum_{x=1}^{w} \sum_{y=1}^{h} ((I)_{x,y} - LMFN(I_{Blur})_{x,y})^2 \tag{4}$$

where $I_{Blur}$ represents the input blurred image, $LMFN$ represents our network, *I* is the standard sharp image, *w* and *h* are the length and width of the input/output image, respectively.

## 4. EXPERIMENTS

### 4.1. Implementation details

Our network is trained using images from GoPro [4] and Kohler datasets [26] without augmentation. We set the number of RFDB layers in our model as 4. For model optimization, we adopt Adam with momentum as 0.9 and weight decay as 1e-4. The learning rate is initialized as 1e-4 and decrease with a factor of 10 for every 5×105 iterations. All experiments were conducted using Pytorch on NVIDIA 2080Ti GPUs. PSNR, SSIM, model size and inference time are adopted to evaluate our method.

### 4.2. Quantitative and qualitative evaluation

**Table 1.** Performance and efficiency comparison on the GoPro dataset.

| Methods | PSNR | SSIM | Model Size | Time |
|---|---|---|---|---|
| DeepDeblur [4] | 29.08 | 0.9135 | 303.6M | 15s |
| Tao et al. [5] | 30.10 | 0.9323 | 33.6M | 1.6s |
| Gao et al. [6] | 30.92 | 0.9421 | 2.84M | 1.6s |
| DMPHN [7] | 30.21 | 0.9345 | 21.7M | 0.03s |
| Zhang et al. [8] | 29.19 | 0.9306 | 37.1M | 1.4s |
| DeblurGAN [10] | 28.70 | 0.927 | 37.1M | 0.85s |
| DeblurGANv2 [11] | 29.55 | 0.934 | 15M | 0.35s |
| SIS [27] | 30.28 | 0.912 | 36.54M | 0.303s |
| Yuan et al. [28] | 29.81 | 0.9368 | 3.1M | 0.01s |
| LMFN(Ours) | **31.54** | 0.923 | **1.25M** | **0.019s** |

In order to prove the effectiveness of our proposed method, we compare the performance of our approach with some state-of-the-art deblurring methods [4,5,6,7,8,10,11,27,28] on the GoPro dataset. The quantitative results are shown in Table 1, and visual comparisons are shown in Fig. 4. As can be seen from Table 1, the number of parameters in DeepDeblur is the largest among all methods, which is because it contains a mass of convolution layers. Through introducing the strategies of parameter sharing [5,6,8,27], GAN [10,11], hierarchical multi-patch [7] and optical flow [28], the number of parameters can be effectively reduced. Since we encode and fuse the hierarchical image information in smaller-scale spaces and adopt the RFBD in decoding stage, the model size of our LMFN is smaller than the other algorithms. Besides, we can also find that the PSNR obtained by our LMFN is superior to other approaches, this may attribute to the attention mechanisms utilized in it. In the visual comparison, it can be seen that most methods fail to recover the clear persons and car from the first and second images with severe motion blur. For the third image, we can still observe some noticeable blur artifacts in the results of some methods (such as DMPHN and SIS). On the contrary, our LMFN can effectively obtain the clearer details and recover sharper images than other methods.

We also report the quantitative results on the Kohler dataset in Table 2. From these results, we can also find that our LMFN outperforms other methods in both deblurring quality and model size. Furthermore, it can be seen from the visual comparison in Fig. 5 that our approach recovers more detailed information from the blurred images.

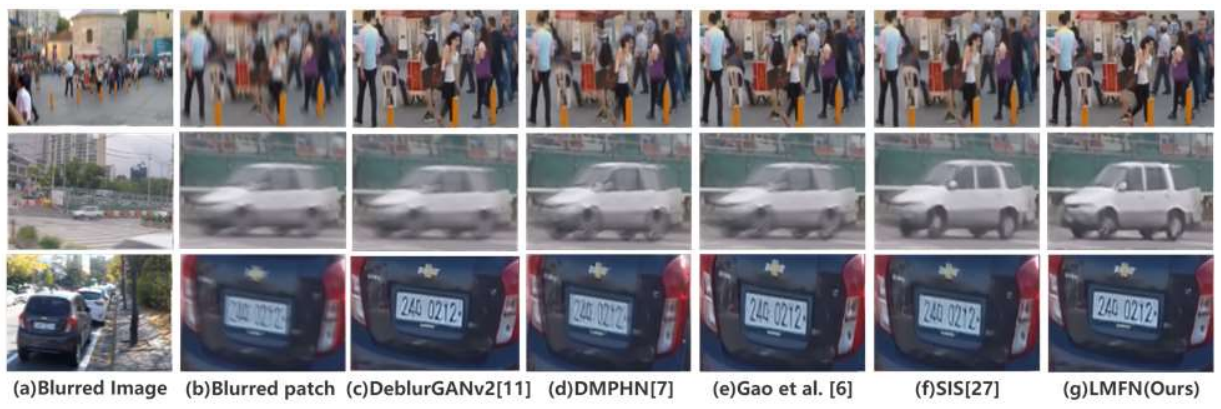

**Fig. 4.** Visual comparison of the deblurring results obtained by some methods on GoPro dataset. Zoom in for better view.

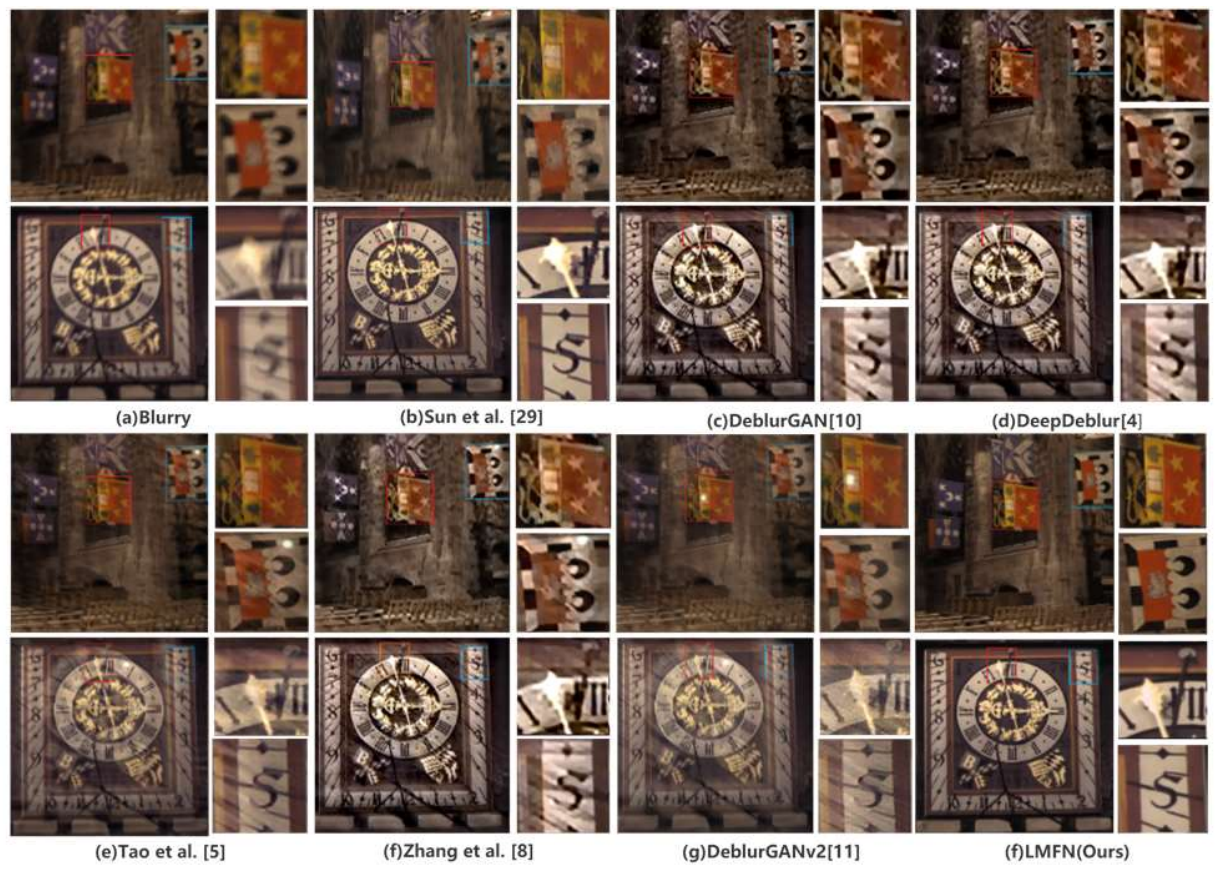

**Fig. 5.** Some visual comparison examples on Kohler dataset. Zoom in for better view.

**Table2.** Performance and efficiency comparison on the Kohler dataset.

| Methods | PSNR | SSIM | Model Size | Time |
|---|---|---|---|---|
| DeepDeblur [4]. | 26.48 | 0.807 | 303.6M | 15s |
| Tao et al. [5] | 26.57 | 0.8373 | 33.6M | 1.6s |
| Zhang et al. [8] | 24.21 | 0.7562 | 37.1M | 1.4s |
| DeblurGAN [10] | 26.10 | 0.807 | 37.1M | 0.85s |
| DeblurGANv2 [11] | 26.97 | 0.830 | 15M | 0.35s |
| Sun et al. [29] | 25.22 | 0.773 | 54.1M | 1,200s |
| Xu et al. [1] | 27.47 | 0.811 | N/A | 13.41s |
| LMFN(Ours) | **30.94** | **0.9027** | **1.25M** | **0.019s** |

### 4.3. Ablation study

In this section, we conduct several experiments to evaluate the effectiveness of each component in our method. First, we replace the multi-scale hierarchical fusion module (MSHF) with traditional convolution layers in our network. The stride of convolution is set as 2 and two residual models are applied to resampling the feature. Then, the distillation blocks in the decoding stage of our approach are removed and the same number of traditional residual blocks are employed for sharp image reconstruction. Finally, we discard the attention based information fusion modules (i.e., the ALFM and ACFM) and employed a standard concatenation to combine the features obtained by different RFDB layers and feature channels of the last RFDB. Through the above settings, we can get three new network structures, i.e., the proposed network LMFN without MSHF, without distillation block and without ALFM /ACFM. In order to fairly compare the proposed model with different structures, we adopted the same parameter setting to train the networks. The results of ablation experiments on the GoPro dataset are summarized in Table 3.

**Table 3.** Quantitative comparison of different ablations of our network on the GroPro dataset.

| MSHF | distillation block | ALFM/ ACFM | PSNR | SSIM | Model Size |
|---|---|---|---|---|---|
| X | √ | √ | 30.8 | 0.917 | 1.8M |
| √ | X | √ | 29.2 | 0.891 | 2.9M |
| √ | √ | X | 31.0 | 0.911 | 1.20M |
| √ | √ | √ | **31.54** | **0.923** | **1.25M** |

As can be seen from the table, The MSHF, distillation block and ALFM/ACFM are all essential for our proposed model. Without MSHF in our model, the value of PSRN decreases by 0.74 and the model size increases by 0. 45. This means that MSHF can extract better features and help to reduce the network parameters. The decrease of deblurring quality is obvious when the distillation block removed. This is because the traditional residual blocks need to build deeper networks to extract the useful information, thus just a few blocks cannot learn image features well. Moreover, replacing distillation block with traditional residual block also apparently increase the model size. At last, we can see that the ALFM and ACFM which fuse the features of different layers and channels by attention mechanism are also indispensable to improve the performance of our network.

Finally, we compare the performances of our LMFN with different numbers of RFDBs. From Table 4, we can find our method performs worse when the number of RFDB layers is small, while too many RFDBs would not further increase the performance. The results also justify that setting the number of RFDB layers as 4 is reasonable in our model.

**Table 4.** Performance comparison of different distillation blocks setting on the GroPro dataset.

| distillation blocks setting | 1 | 2 | 3 | 4 | 5 | 6 |
|---|---|---|---|---|---|---|
| **PSNR** | 29.8 | 29.7 | 30.7 | 31.54 | 31.52 | 31.54 |
| **Model Size(M)** | 1.15 | 1.18 | 1.22 | 1.25 | 1.28 | 1.31 |

## 5. CONCLUSION

In this work, we propose a lightweight multi-information fusion network (LMFN) for image deblurring tasks. In order to make the network have fewer parameters and faster speed, the multi-scale hierarchical fusion module and residual feature distillation block are adopted in the encoding and decoding stages respectively. Moreover, two attention based modules are also proposed to improve the feature representation power in our approach. A large number of experiments on two datasets show that the LMFN achieves good results in both image deblurring quality and model complexity.